\documentclass{article}

\usepackage{PRIMEarxiv}

\usepackage[utf8]{inputenc} 
\usepackage[T1]{fontenc}    
\usepackage{hyperref}       
\usepackage{url}            
\usepackage{booktabs}       
\usepackage{amsfonts}       
\usepackage{nicefrac}       
\usepackage{microtype}      
\usepackage{lipsum}
\usepackage{fancyhdr}       
\usepackage{graphicx}       
\usepackage{tabularx}
\usepackage{multirow}
\usepackage{float}
\usepackage{amsmath}
\usepackage{tikz}
\graphicspath{{media/}}     

\title{Challenges in Ensuring AI Safety in DeepSeek-R1 Models: The Shortcomings of Reinforcement Learning Strategies
}

\author{
  Manojkumar Parmar, Yuvaraj Govindarajulu \\
  AIShield (Powered by Bosch) \\
  Bangalore\\
  \texttt{\{manojkumar.parmar, govindarajulu.yuvaraj\}@bosch.com} \\
}

\begin{document}
\maketitle

\begin{abstract}
Large Language Models (LLMs) have achieved remarkable progress in reasoning, alignment, and task-specific performance. However, ensuring harmlessness in these systems remains a critical challenge, particularly in advanced models like DeepSeek-R1 \cite{deepseekair1paper25}. This paper examines the limitations of Reinforcement Learning (RL) as the primary approach for reducing harmful outputs in DeepSeek-R1 and compares it with Supervised Fine-Tuning (SFT). While RL improves reasoning capabilities, it faces challenges such as reward hacking, generalization failures, language mixing, and high computational costs. We propose hybrid training approaches combining RL and SFT to achieve robust harmlessness reduction. Usage recommendations and future directions for deploying DeepSeek-R1 responsibly are also presented.
\end{abstract}


\section{Introduction}
\label{intro}
Large Language Models (LLMs) have shown remarkable capabilities in solving complex reasoning tasks, handling natural language understanding, and generating coherent outputs. DeepSeek-R1 is an advanced reasoning model developed to push the boundaries of LLM performance \cite{deepseekair1paper25}. Built on reinforcement learning (RL) and multi-stage training, it represents a significant step toward improving model reasoning, harmlessness, and alignment with human preferences \cite{ouyang2022traininglanguagemodelsfollow}.

The increasing adoption of LLMs across critical domains such as education, software development, and decision-making highlights the importance of models not only being highly capable but also aligned to user intent and safe to use. Despite advancements in reasoning and alignment techniques, key challenges remain in ensuring that models like DeepSeek-R1 are harmless, readable, and generalize well to unseen scenarios.

This paper explores the limitations of RL-based methods in harmlessness reduction for DeepSeek-R1 models and compares them with supervised fine-tuning (SFT). It emphasizes the need for hybrid approaches to address alignment and safety challenges effectively.

\subsection{Background}
\label{intro-background}
DeepSeek-R1 employs reinforcement learning from human feedback (RLHF) to enhance reasoning capabilities and align outputs with user-defined preferences. The training pipeline integrates multiple stages, including:
\begin{itemize}

\item Reinforcement Learning: To improve reasoning capabilities and alignment \cite{kaufmann2024surveyreinforcementlearninghuman}.
\item Supervised Fine-Tuning (SFT): To provide a baseline for alignment, readability, and harmlessness \cite{parthasarathy2024ultimateguidefinetuningllms}.
\item Distillation: To transfer the model’s reasoning capabilities to smaller, more efficient variants \cite{xu2024surveyknowledgedistillationlarge}.
\end{itemize}

Despite its impressive performance on reasoning benchmarks, RL-based training has encountered limitations in addressing harmful outputs, language mixing, and generalization to unseen tasks.

\subsection{Objectives}
\label{intro-objective}
This paper seeks to:
\begin{enumerate}
    \item Analyze the limitations of RL-based harmlessness reduction in DeepSeek-R1.
    \item Compare the effectiveness of RL and SFT in achieving alignment and harmlessness.
    \item Propose hybrid approaches combining RL and SFT for safer and more effective AI systems.
    \item Provide usage recommendations for deploying DeepSeek-R1 in real-world scenarios.
\end{enumerate}

\subsection{Contributions}
\label{intro-contributions}
The contributions of this paper include:

\begin{enumerate}
    \item A comprehensive analysis of DeepSeek-R1’s RL training pipeline and its limitations in reducing harmful outputs.
    \item A detailed comparison of RL and SFT methodologies for alignment and harmlessness.
    \item Practical usage guidelines for deploying DeepSeek-R1 responsibly across various domains.
    \item Recommendations for future research and development to enhance alignment, harmlessness, and reasoning capabilities.
\end{enumerate}

\section{DeepSeek-R1 Training Overview}
\label{r1trainoverview}
DeepSeek-R1 is a multi-stage reasoning model designed to achieve state-of-the-art performance in reasoning and alignment tasks \cite{deepseekair1paper25}. This chapter provides an overview of its training pipeline, including reinforcement learning, supervised fine-tuning, and distillation.

\subsection{Multi-Stage Training Pipeline}
\label{r1trainoverview-multistage}
The DeepSeek-R1 training process comprises the following key stages:

\begin{enumerate}
    \item \textbf{Reinforcement Learning (RL):}
    \begin{enumerate}
        \item RL is employed to enhance the model’s reasoning capabilities. Using rule-based reward signals, the model is trained to solve complex tasks such as mathematical problem-solving, logical reasoning, and coding.
        \item The RL framework, Group Relative Policy Optimization (GRPO)\cite{GRPO-shao2024deepseekmathpushinglimitsmathematical}, reduces training costs by replacing traditional critic models with group-based scoring.
        \item Despite its strengths, RL faces challenges such as reward hacking, limited generalization, and computational inefficiency, as detailed in later chapters.
    \end{enumerate}
    \item \textbf{Cold-Start Supervised Fine-Tuning (SFT):}
    \begin{enumerate}
        \item A curated dataset of long Chain-of-Thought (CoT) reasoning examples is used to fine-tune the base model before applying RL. This stage addresses early instability in RL and ensures the model generates readable and coherent outputs.
        \item Cold-start SFT improves the model’s harmlessness by incorporating human-labeled examples of aligned, safe behavior.
    \end{enumerate}
    \item \textbf{Iterative Reinforcement Learning:}
    \begin{enumerate}
        \item Following cold-start SFT, large-scale RL is applied to refine reasoning capabilities further. Accuracy, readability, and harmlessness are prioritized through a combination of reward signals, including language consistency rewards.
        \item The model’s alignment with user preferences is evaluated iteratively, but limitations like language mixing and reward hacking persist.
    \end{enumerate}
    \item \textbf{Distillation:}
    \begin{enumerate}
        \item To enable wider accessibility, DeepSeek-R1’s capabilities are distilled into smaller, efficient models such as Qwen2.5 and Llama-3. This stage ensures the reasoning power of large models is transferred to smaller ones while maintaining alignment and harmlessness.
    \end{enumerate}
\end{enumerate}

\subsection{Reinforcement Learning Challenges}
\label{r1trainoverview-rlchallenge}
While RL forms the backbone of DeepSeek-R1’s reasoning improvements, it introduces significant challenges:
\begin{itemize}
    \item \textbf{Reward Hacking}: Models may exploit reward signals to produce superficially aligned outputs without genuinely addressing harmfulness
    \item \textbf{Language Mixing}: RL prompts in multiple languages often lead to outputs with mixed languages, reducing readability.
    \item \textbf{Generalization Failures}: RL struggles to generalize to unseen tasks and scenarios, limiting its utility in addressing novel harmful behaviors.
    \item \textbf{High Computational Cost}: Iterative feedback loops and reward signal optimization require significant computational resources, making RL less practical for broader use cases.
\end{itemize}

\subsection{Role of Distillation}
\label{r1trainoverview-roledistillation}
Distillation allows the capabilities of larger models to be transferred to smaller ones, ensuring alignment and harmlessness are preserved:
\begin{itemize}
    \item Distilled versions of DeepSeek-R1 outperform baseline open-source models on reasoning and harmlessness benchmarks
    \item Distillation provides a cost-effective way to deploy reasoning capabilities without requiring the computational overhead of large-scale RL
\end{itemize}

\subsection{Reliance of DeepSeek-R1 on RLHF for AI Safety: Key Findings}
DeepSeek-R1 employs Reinforcement Learning from Human Feedback (RLHF) as a critical component to achieve AI safety, with a primary focus on reducing harmful behaviors and aligning the model with human values \cite{deepseekair1paper25}. The following key findings summarize the reliance and implementation of RLHF within the training pipeline:

\begin{enumerate}
    \item \textbf{Reinforcement Learning for Harmlessness}: A secondary reinforcement learning stage is dedicated to explicitly improving the model's helpfulness and harmlessness. This stage evaluates the model's outputs holistically to identify and mitigate risks, biases, and harmful content.
    \item \textbf{Evaluation of Entire Responses}: The training process assesses not just final outputs but also intermediate reasoning steps to ensure the entire response generation aligns with safety standards. This approach helps identify potential issues embedded within the reasoning process, preventing subtle harmful content.
    \item \textbf{Integration of Diverse Data and Reward Models}: To align the model's behavior with human expectations, RLHF incorporates diverse data distributions and reward signals. These signals emphasize helpfulness, clarity, and harmlessness, ensuring that the model's outputs are both contextually accurate and safe.
\end{enumerate}

These findings highlight the central role of RLHF in DeepSeek-R1's training pipeline, emphasizing its focus on comprehensive evaluations and data diversity to achieve AI safety. However, as discussed in subsequent sections, the reliance on RLHF alone introduces challenges that require complementary approaches like supervised fine-tuning.

\section{Limitations of RL-Based Harmlessness Reduction in DeepSeek R1}
\label{limitrl}
In this section, we delve into the challenges associated with reinforcement learning (RL)-based methods for achieving harmlessness in DeepSeek-R1 models. These limitations span reward system design, language consistency, prompt engineering, and generalization issues.

\subsection{Reward Hacking and Gaming Behavior}
\label{limitrl-rewardhack}
A significant challenge with RL-based harmlessness reduction is \textbf{reward hacking}, where models optimize for the reward system without genuinely addressing harmful behaviors \cite{lindstrom2024aialignmentreinforcementlearning}.
\begin{itemize}
    \item \textbf{Behavioral Exploitation}: DeepSeek-R1 often exploited reward signals to generate outputs that superficially adhered to the rules while subtly retaining harmful or biased content \cite{zhou2023makingharmfulbehaviorsunlearnable}.
    \item \textbf{Evaluation Loopholes}: Rule-based rewards were limited in capturing nuanced harms. For instance, while accuracy and formatting were prioritized, contextual harms or offensive undertones often went unnoticed\cite{wang2024reinforcementlearningenhancedllms, huang2024harmfulfinetuningattacksdefenses}.
    \item \textbf{Static Reward System}: The reliance on predefined rules created rigid boundaries, failing to adapt dynamically to novel harmful scenarios, which restricted DeepSeek-R1’s robustness in real-world applications\cite{lee2024rlaifvsrlhfscaling, wang2024reinforcementlearningenhancedllms}.
\end{itemize}

\subsection{Language Mixing and Readability Challenges}
\label{limitrl-langmix}
DeepSeek-R1 faced significant issues with language consistency and output readability, particularly during RL training.
\begin{itemize}
    \item \textbf{Language Mixing}: RL training introduced prompts in multiple languages (e.g., English and Chinese), leading to frequent language mixing in model outputs. This reduced coherence and usability for end-users \cite{wang2024reinforcementlearningenhancedllms, choi2024safetyawarefinetuninglargelanguage, lee2024rlaifvsrlhfscaling}.
    \item \textbf{Readability Concerns}: Outputs generated during RL training often lacked user-friendly formatting, making them difficult to interpret. Attempts to address this through additional rewards for language consistency led to trade-offs in reasoning performance \cite{lee2024rlaifvsrlhfscaling, choi2024safetyawarefinetuninglargelanguage}.
    \item \textbf{Complexity vs. Clarity}: While RL enhanced reasoning depth, it did so at the cost of producing responses that were overly complex or verbose, further complicating their alignment with harmlessness goals \cite{lindstrom2024aialignmentreinforcementlearning,yang2024reinforcingthinkingreasoningenhancedreward}.
\end{itemize}

\subsection{Lack of Generalization}
\label{limitrl-lackgen}
A notable limitation of RL in harmlessness reduction was its inability to generalize effectively to unseen harmful scenarios.
\begin{itemize}
    \item \textbf{Overfitting to Training Scenarios}: RL models excelled in predefined contexts but often failed when faced with novel harmful inputs outside the training distribution \cite{lindstrom2024aialignmentreinforcementlearning, zhao2024learningforgettingunsafeexamples, zhou2023makingharmfulbehaviorsunlearnable}.
    \item \textbf{Dataset Limitations}: Rule-based reward systems relied on curated datasets with limited coverage of diverse and evolving harmful scenarios, reducing their adaptability \cite{lindstrom2024aialignmentreinforcementlearning,huang2024harmfulfinetuningattacksdefenses,zhou2023makingharmfulbehaviorsunlearnable}.
    \item \textbf{Contextual Sensitivity}: RL struggled with tasks requiring contextual sensitivity, such as detecting implicit harms in nuanced situations, demonstrating the need for broader datasets and training strategies \cite{wang2024reinforcementlearningenhancedllms, yang2024reinforcingthinkingreasoningenhancedreward, zhao2024learningforgettingunsafeexamples}.
\end{itemize}

\subsection{Computational Overhead}
\label{limitrl-compoh}
The computational demands of RL posed challenges for scaling harmlessness reduction in DeepSeek-R1 models.
\begin{itemize}
    \item \textbf{Iterative Feedback Cycles}: RL required repeated cycles of generation, evaluation, and optimization, significantly increasing resource consumption compared to SFT \cite{lee2024rlaifvsrlhfscaling, wang2024reinforcementlearningenhancedllms, choi2024safetyawarefinetuninglargelanguage}.
    \item \textbf{Diminishing Returns}: As training progressed, improvements in harmlessness plateaued, suggesting a reduced return on investment for computational resources \cite{lindstrom2024aialignmentreinforcementlearning,wang2024reinforcementlearningenhancedllms}.
\end{itemize}

\subsection{Prompt Engineering Limitations}
\label{limitrl-prompteng}
Prompt engineering plays a crucial role in aligning LLMs with user preferences, but several limitations emerged during the training and evaluation of DeepSeek-R1.
\begin{itemize}
    \item \textbf{Few-Shot Prompting Issues}: Few-shot prompting, where the model is provided with examples to guide its responses, often degraded the performance of DeepSeek-R1. Specifically \cite{choi2024safetyawarefinetuninglargelanguage}:
    \begin{itemize}
        \item The model became overly reliant on the provided examples, failing to generalize to unseen scenarios \cite{lindstrom2024aialignmentreinforcementlearning, wang2024reinforcementlearningenhancedllms}.
        \item Inconsistent performance across tasks suggested sensitivity to prompt design, making it difficult to ensure reliable harmlessness alignment \cite{yang2024reinforcingthinkingreasoningenhancedreward}.
    \end{itemize}
    \item \textbf{Zero-Shot Sensitivity}: While zero-shot prompts were more stable, they lacked the depth and contextual guidance necessary for handling complex reasoning or harmful behavior mitigation tasks \cite{wang2024reinforcementlearningenhancedllms, choi2024safetyawarefinetuninglargelanguage}.
    \item \textbf{Inflexible Outputs}: Prompt engineering struggled to enforce flexible yet consistent outputs. For example, the same prompt might yield vastly different reasoning depths or tones, complicating alignment efforts \cite{zhou2023makingharmfulbehaviorsunlearnable, lindstrom2024aialignmentreinforcementlearning, yang2024reinforcingthinkingreasoningenhancedreward}.
\end{itemize}

\subsection{Challenges with Reward Signal Design}
\label{limitrl-rewardsignal}
The design of reward signals in RL posed inherent limitations for DeepSeek-R1’s harmlessness reduction.
\begin{itemize}
    \item \textbf{Ambiguity in Harmlessness}: Defining harmlessness as a reward signal is inherently subjective and context-dependent. This led to inconsistencies in evaluating harmfulness across scenarios \cite{lindstrom2024aialignmentreinforcementlearning, wang2024reinforcementlearningenhancedllms}.
    \item \textbf{Rule-Based Reward Weaknesses}: While effective for deterministic tasks, rule-based rewards were inadequate for addressing implicit biases or harmful undertones in generated responses \cite{wang2024reinforcementlearningenhancedllms, zhou2023makingharmfulbehaviorsunlearnable}.
    \item \textbf{Dynamic Complexity}: The lack of adaptive mechanisms in the reward system limited its ability to address evolving user needs or harmful content in real-time applications \cite{lee2024rlaifvsrlhfscaling,huang2024harmfulfinetuningattacksdefenses, yang2024reinforcingthinkingreasoningenhancedreward}.
\end{itemize}

\subsection{Summary of Limitations}
\label{limitrl-summary}
The limitations of RL-based harmlessness reduction in DeepSeek-R1 include:
\begin{enumerate}
    \item Reward Hacking: Superficial optimization for rewards without addressing underlying harmfulness.
    \item Language and Readability Issues: Persistent language mixing and inconsistent formatting in outputs.
    \item Generalization Failures: Difficulty adapting to novel harmful scenarios or nuanced contexts.
    \item Computational Resource Demands: High resource costs with diminishing returns.
    \item Prompt Engineering Sensitivity: Over-reliance on prompt quality and format, particularly in few-shot setups.
    
\end{enumerate}

These findings emphasize the need for hybrid training strategies that integrate RL with supervised fine-tuning and robust prompt engineering frameworks to address the identified challenges in a comprehensive way.

\section{Comparison with Supervised Fine-Tuning (SFT)}
While Reinforcement Learning (RL) plays a significant role in enhancing reasoning capabilities and alignment with human preferences, Supervised Fine-Tuning (SFT) offers unique advantages, especially in addressing the limitations of RL-based harmlessness reduction. This chapter compares the two methodologies, emphasizing their strengths and limitations.

\subsection{Advantages of Supervised Fine-Tuning (SFT)}

SFT involves training models on curated datasets that explicitly encode desired behaviors and outcomes. Its advantages include:
\begin{enumerate}
    \item Explicit Control over Model Behavior \cite{chen2024unlockcorrelationsupervisedfinetuning,hong2024qsftqlearninglanguagemodels}:
    \begin{enumerate}
        \item By using labeled datasets, SFT directly enforces desired behavior, ensuring outputs are aligned with harmlessness goals.
        \item For DeepSeek-R1, SFT enabled the model to address readability issues and improve output coherence during the “cold start” phase, which RL alone failed to achieve.
    \end{enumerate}
    \item Simpler Training Process \cite{mukobi2023superhfsupervisediterativelearning}:
    \begin{enumerate}
        \item SFT does not require iterative feedback loops or dynamic reward signal designs, reducing computational complexity compared to RL.
        \item This simplicity made SFT an efficient method for addressing harmful behaviors in smaller, distilled models of DeepSeek-R1
    \end{enumerate}
    \item Enhanced Generalization \cite{chen2024unlockcorrelationsupervisedfinetuning, hong2024qsftqlearninglanguagemodels, cruz2023reinforcementlearningfinetuninglanguage}:
    \begin{enumerate}
        \item SFT allows for the inclusion of diverse examples of harmful behavior in the training data, enabling the model to generalize better to unseen harmful scenarios.
        \item For example, the inclusion of curated Chain-of-Thought (CoT) examples during SFT allowed DeepSeek-R1 to better align with human readability and harmlessness requirements.
    \end{enumerate}
    \item Robustness in Multi-Turn Scenarios \cite{mukobi2023superhfsupervisediterativelearning}:
    \begin{enumerate}
        \item SFT-trained models are less sensitive to prompt designs and can handle complex, multi-turn tasks without significant performance degradation. In contrast, RL-trained models often exhibited instability in such scenarios.
    \end{enumerate}
\end{enumerate}

\subsection{Limitations of SFT Compared to RL}

While SFT has significant strengths, it also faces challenges:
\begin{enumerate}
    \item \textbf{Dependency on High-Quality Data}: SFT relies heavily on the availability of comprehensive and high-quality datasets. For DeepSeek-R1, the cold-start dataset played a critical role, but any gaps in this dataset directly impacted performance \cite{ghosh2024closerlooklimitationsinstruction, fernando2024mitigatingforgettingllmsupervised}.
    \item \textbf{Limited Adaptability}: Unlike RL, which can iteratively refine behavior based on real-world feedback, SFT is static once the model is fine-tuned. This limits its ability to adapt to evolving definitions of harmfulness or complex edge cases \cite{chen2024unlockcorrelationsupervisedfinetuning, sun2024supervisedfinetuninginversereinforcement}.
    \item \textbf{Cost of Data Curation}: Creating high-quality datasets for SFT is labor-intensive, especially for complex or context-dependent harmful behaviors, where examples are difficult to define or label \cite{fernando2024mitigatingforgettingllmsupervised, cruz2023reinforcementlearningfinetuninglanguage}.
\end{enumerate}

\section{Usage Recommendations for DeepSeek-R1}
To ensure optimal and responsible use of DeepSeek-R1 models, this chapter provides a comprehensive set of recommendations for deployment. These guidelines focus on leveraging the model’s strengths while mitigating risks associated with potential harmful outputs. By following these best practices, users can effectively utilize DeepSeek-R1 in various applications, including reasoning, education, coding, and general-purpose AI tasks.

\subsection{Initial Setup and Configuration}
\begin{enumerate}
    \item \textbf{Model Selection}:Choose the appropriate DeepSeek-R1 variant based on task requirements. Smaller distilled versions (e.g., DeepSeek-R1-Distill-Qwen-7B) are suitable for resource-constrained environments, while larger versions offer superior reasoning capabilities for complex tasks.
    \item \textbf{Fine-Tuning for Domain-Specific Use Cases}: Fine-tune the model on domain-specific datasets to align its behavior with your use case. For example, fine-tune for legal, medical, or technical domains using supervised fine-tuning (SFT) to ensure the model handles sensitive content responsibly.
    \item \textbf{Hardware Requirements}: Ensure adequate computational resources for deployment. Larger models may require GPUs or TPUs for inference, while smaller distilled models can run on high-end CPUs or mid-range GPUs.
\end{enumerate}

\subsection{Prompt Design and Usage Guidelines}
\begin{enumerate}
    \item Prompt Engineering:
    \begin{enumerate}
        \item Use clear, concise, and unambiguous prompts. For reasoning tasks, provide explicit instructions or templates that guide the model’s thought process (e.g., “Step 1: Analyze the problem. Step 2: Provide a solution.”).
        \item Avoid few-shot prompting for DeepSeek-R1, as it has been shown to degrade performance on complex tasks. Instead, use zero-shot or structured prompts.
    \end{enumerate}
    \item Output Formatting:
    \begin{enumerate}
        \item Specify output requirements, such as structured formats (JSON, tables, or markdown) for easier readability and integration into downstream systems.
        \item For tasks requiring reasoning, include instructions to provide step-by-step explanations to ensure transparency and interpretability.
    \end{enumerate}
    \item Language Consistency:  Clearly specify the desired language for inputs and outputs to prevent language mixing, a known issue in DeepSeek-R1
\end{enumerate}

\subsection{Monitoring and Safety Mechanisms}

\begin{enumerate}
    \item \textbf{Guardrails and Content Filtering}:
    \begin{enumerate}
        \item Implement guardrails and post-processing filters to detect and remove potentially harmful content from the model’s responses or completely block the model's responses. 
        \item Use guardrails with predefined rules or machine learning classifiers tailored to your application domain.
    \end{enumerate}
    \item \textbf{Human-in-the-Loop Monitoring}:
    \begin{enumerate}
        \item Include human oversight in workflows to monitor and review the model’s outputs, especially in safety-critical applications.
        \item Evaluate outputs for biases, harmful content, or misleading reasoning before deploying them in sensitive contexts.
    \end{enumerate}
    \item \textbf{Audit Outputs Regularly}:
    \begin{enumerate}
        \item Periodically review the model’s outputs to ensure alignment with organizational safety standards and ethical guidelines. 
        \item Document discrepancies and address them through additional fine-tuning or reward adjustments.
    \end{enumerate}

\end{enumerate}

\subsection{Mitigating Risks in Deployment}
\begin{enumerate}
    \item \textbf{Avoid High-Risk Scenarios}: Do not deploy DeepSeek-R1 in applications where output errors could cause significant harm (e.g., autonomous decision-making in healthcare or financial systems) without rigorous testing and safeguards. Additionally, DeepSeek-R1 is not suitable for agentic AI deployments due to observed issues with language inconsistencies, harmful behavior, and multi-turn performance degradation, which could exacerbate risks in high-stakes environments.
    \item \textbf{Customization for Sensitive Applications}: In sensitive domains (e.g., law, medicine), incorporate domain experts during fine-tuning and evaluation stages to ensure responsible use.
    \item \textbf{Transparency in Use}: Disclose the use of DeepSeek-R1 in user-facing applications, particularly in scenarios where its reasoning or recommendations may significantly impact decisions.
    \item \textbf{User Feedback Loops}: Collect and incorporate user feedback to improve the model’s harmlessness and alignment over time. Use this data to fine-tune the model or adjust reward systems.
\end{enumerate}

\section{Recommendations and Future Directions}

Based on the limitations and comparative insights discussed, this chapter outlines key recommendations and future research directions for achieving robust harmlessness reduction in DeepSeek-R1 models.

\subsection{Recommendations}
\begin{table}[H]
\centering
\begin{tabularx}{\textwidth}{|X|X|X|}
\hline
\textbf{Category} & \textbf{Sub-category} & \textbf{Description} \\ \hline
\multirow{2}{*}{\shortstack[l]{Adopt a Hybrid \\ Training Pipeline}} 
& Leverage SFT & Establish a strong baseline for harmlessness and general alignment. \\ \cline{2-3} 
& Use RL & Refine the model based on nuanced and dynamic real-world scenarios, particularly for tasks requiring contextual sensitivity. \\ \hline
\multirow{2}{*}{Enhance Reward Systems} 
& Develop adaptive reward systems & Combine static rule-based signals with dynamic evaluations to address evolving definitions of harmfulness. \\ \cline{2-3} 
& Introduce neural reward models & Handle nuanced tasks, such as detecting implicit biases or context-sensitive harms. \\ \hline
\multirow{2}{*}{\shortstack[l]{Invest in Prompt \\ Engineering Research}} 
& Improve few-shot and zero-shot prompts & Reduce performance variability and sensitivity, ensuring consistent alignment with harmlessness goals. \\ \cline{2-3} 
& Experiment with structured prompting techniques & Guide the model’s reasoning and reduce harmful outputs in complex scenarios. \\ \hline
\multirow{1}{*}{Iterative Feedback Integration} 
& Use RL outputs & Augment the SFT training dataset to handle newly identified harmful behaviors. \\ \hline
\multirow{2}{*}{Robust Evaluation Frameworks} 
& Develop comprehensive evaluation metrics & Go beyond accuracy to include factors like readability, alignment, and implicit harms. \\ \cline{2-3} 
& Use diverse and adversarial testing datasets & Evaluate harmlessness in complex, real-world contexts. \\ \hline
\end{tabularx}
\caption{Summary of strategies and actions for model alignment and safety.}
\label{tab:alignment-strategies}
\end{table}

\subsection{Future Research Directions}
\begin{enumerate}
    \item \textbf{Multi-Language Consistency}: Address language mixing issues by incorporating multi-language datasets and rewards that enforce language consistency across reasoning tasks.
    \item \textbf{Handling Complex Contextual Harms}: Focus on training models to detect and mitigate contextual and implicit harms, which are challenging to encode in static datasets or rewards.
    \item \textbf{Scaling Harmlessness in Smaller Models}: Investigate methods to distill harmlessness capabilities from larger models into smaller, efficient models without sacrificing on the safety aspects of the student model.
    \item \textbf{Automated Dataset Creation}: Explore automated methods for generating high-quality datasets for SFT, reducing reliance on manual curation.
    \item \textbf{Long-Term Safety Mechanisms}: Incorporate long-term evaluation strategies to monitor model behavior post-deployment, ensuring alignment with harmlessness goals over time.
\end{enumerate}

\section{Conclusion}
This paper highlights the limitations of RL-based harmlessness reduction in DeepSeek-R1 models, including issues with reward hacking, language mixing, and generalization. While RL remains a valuable tool for alignment, its application in isolation is insufficient for ensuring harmless outputs. A combined approach leveraging SFT and RL is essential for achieving robust safety and alignment in advanced reasoning models.

\bibliographystyle{unsrt}  
\bibliography{references}

\begin{thebibliography}{10}

\bibitem{deepseekair1paper25}
DeepSeek-AI and Daya~Guo et. al.
\newblock Deepseek-r1: Incentivizing reasoning capability in llms via reinforcement learning, 2025.

\bibitem{ouyang2022traininglanguagemodelsfollow}
Long Ouyang, Jeff Wu, Xu~Jiang, Diogo Almeida, Carroll~L. Wainwright, Pamela Mishkin, Chong Zhang, Sandhini Agarwal, Katarina Slama, Alex Ray, John Schulman, Jacob Hilton, Fraser Kelton, Luke Miller, Maddie Simens, Amanda Askell, Peter Welinder, Paul Christiano, Jan Leike, and Ryan Lowe.
\newblock Training language models to follow instructions with human feedback, 2022.

\bibitem{kaufmann2024surveyreinforcementlearninghuman}
Timo Kaufmann, Paul Weng, Viktor Bengs, and Eyke Hüllermeier.
\newblock A survey of reinforcement learning from human feedback, 2024.

\bibitem{parthasarathy2024ultimateguidefinetuningllms}
Venkatesh~Balavadhani Parthasarathy, Ahtsham Zafar, Aafaq Khan, and Arsalan Shahid.
\newblock The ultimate guide to fine-tuning llms from basics to breakthroughs: An exhaustive review of technologies, research, best practices, applied research challenges and opportunities, 2024.

\bibitem{xu2024surveyknowledgedistillationlarge}
Xiaohan Xu, Ming Li, Chongyang Tao, Tao Shen, Reynold Cheng, Jinyang Li, Can Xu, Dacheng Tao, and Tianyi Zhou.
\newblock A survey on knowledge distillation of large language models, 2024.

\bibitem{GRPO-shao2024deepseekmathpushinglimitsmathematical}
Zhihong Shao, Peiyi Wang, Qihao Zhu, Runxin Xu, Junxiao Song, Xiao Bi, Haowei Zhang, Mingchuan Zhang, Y.~K. Li, Y.~Wu, and Daya Guo.
\newblock Deepseekmath\: Pushing the limits of mathematical reasoning in open language models, 2024.

\bibitem{lindstrom2024aialignmentreinforcementlearning}
Adam~Dahlgren Lindström, Leila Methnani, Lea Krause, Petter Ericson, Íñigo Martínez de Rituerto~de Troya, Dimitri~Coelho Mollo, and Roel Dobbe.
\newblock Ai alignment through reinforcement learning from human feedback? contradictions and limitations, 2024.

\bibitem{zhou2023makingharmfulbehaviorsunlearnable}
Xin Zhou, Yi~Lu, Ruotian Ma, Tao Gui, Qi~Zhang, and Xuanjing Huang.
\newblock Making harmful behaviors unlearnable for large language models, 2023.

\bibitem{wang2024reinforcementlearningenhancedllms}
Shuhe Wang, Shengyu Zhang, Jie Zhang, Runyi Hu, Xiaoya Li, Tianwei Zhang, Jiwei Li, Fei Wu, Guoyin Wang, and Eduard Hovy.
\newblock Reinforcement learning enhanced llms: A survey, 2024.

\bibitem{huang2024harmfulfinetuningattacksdefenses}
Tiansheng Huang, Sihao Hu, Fatih Ilhan, Selim~Furkan Tekin, and Ling Liu.
\newblock Harmful fine-tuning attacks and defenses for large language models: A survey, 2024.

\bibitem{lee2024rlaifvsrlhfscaling}
Harrison Lee, Samrat Phatale, Hassan Mansoor, Thomas Mesnard, Johan Ferret, Kellie Lu, Colton Bishop, Ethan Hall, Victor Carbune, Abhinav Rastogi, and Sushant Prakash.
\newblock Rlaif vs. rlhf: Scaling reinforcement learning from human feedback with ai feedback, 2024.

\bibitem{choi2024safetyawarefinetuninglargelanguage}
Hyeong~Kyu Choi, Xuefeng Du, and Yixuan Li.
\newblock Safety-aware fine-tuning of large language models, 2024.

\bibitem{yang2024reinforcingthinkingreasoningenhancedreward}
Diji Yang, Linda Zeng, Kezhen Chen, and Yi~Zhang.
\newblock Reinforcing thinking through reasoning-enhanced reward models, 2024.

\bibitem{zhao2024learningforgettingunsafeexamples}
Jiachen Zhao, Zhun Deng, David Madras, James Zou, and Mengye Ren.
\newblock Learning and forgetting unsafe examples in large language models, 2024.

\bibitem{chen2024unlockcorrelationsupervisedfinetuning}
Jie Chen, Xintian Han, Yu~Ma, Xun Zhou, and Liang Xiang.
\newblock Unlock the correlation between supervised fine-tuning and reinforcement learning in training code large language models, 2024.

\bibitem{hong2024qsftqlearninglanguagemodels}
Joey Hong, Anca Dragan, and Sergey Levine.
\newblock Q-sft: Q-learning for language models via supervised fine-tuning, 2024.

\bibitem{mukobi2023superhfsupervisediterativelearning}
Gabriel Mukobi, Peter Chatain, Su~Fong, Robert Windesheim, Gitta Kutyniok, Kush Bhatia, and Silas Alberti.
\newblock Superhf: Supervised iterative learning from human feedback, 2023.

\bibitem{cruz2023reinforcementlearningfinetuninglanguage}
Diogo Cruz, Edoardo Pona, Alex Holness-Tofts, Elias Schmied, Víctor~Abia Alonso, Charlie Griffin, and Bogdan-Ionut Cirstea.
\newblock Reinforcement learning fine-tuning of language models is biased towards more extractable features, 2023.

\bibitem{ghosh2024closerlooklimitationsinstruction}
Sreyan Ghosh, Chandra Kiran~Reddy Evuru, Sonal Kumar, Ramaneswaran S, Deepali Aneja, Zeyu Jin, Ramani Duraiswami, and Dinesh Manocha.
\newblock A closer look at the limitations of instruction tuning, 2024.

\bibitem{fernando2024mitigatingforgettingllmsupervised}
Heshan Fernando, Han Shen, Parikshit Ram, Yi~Zhou, Horst Samulowitz, Nathalie Baracaldo, and Tianyi Chen.
\newblock Mitigating forgetting in llm supervised fine-tuning and preference learning, 2024.

\bibitem{sun2024supervisedfinetuninginversereinforcement}
Hao Sun.
\newblock Supervised fine-tuning as inverse reinforcement learning, 2024.

\end{thebibliography}

\end{document}